\documentclass[conference]{IEEEtran}
\IEEEoverridecommandlockouts
\usepackage{cite}
\usepackage{amsmath,amssymb,amsfonts}
\usepackage{algorithmic}
\usepackage{graphicx}
\usepackage{textcomp}
\usepackage{dirtytalk}
\usepackage{xcolor}
\usepackage{booktabs} 
\def\BibTeX{{\rm B\kern-.05em{\sc i\kern-.025em b}\kern-.08em
    T\kern-.1667em\lower.7ex\hbox{E}\kern-.125emX}}
\begin{document}

\title{Inference Optimizations for Large Language Models: Effects, Challenges, and Practical Considerations\\
}

\author{
\IEEEauthorblockN{Donisch, Leo}
\IEEEauthorblockA{\textit{Business Faculty} \\
\textit{Ansbach University of Applied Sciences}\\
Ansbach, Germany \\
donisch19473@hs-ansbach.de}

\and
\IEEEauthorblockN{2\textsuperscript{nd} Sigurd Schacht}
\IEEEauthorblockA{\textit{Faculty of Business} \\
\textit{University of Applied Science}\\
Ansbach, Germany \\
sigurd.schacht@hs-ansbach.de}
\and
\IEEEauthorblockN{3\textsuperscript{rd} Carsten Lanquillon}
\IEEEauthorblockA{\textit{Faculty of Business} \\
\textit{University of Applied Science}\\
Heilbronn, Germany \\
carsten.lanquillon@hs-heilbronn.de}
}

\maketitle
\begin{abstract}
Large language models are ubiquitous in natural language processing because they can adapt to new tasks without retraining. However, their sheer scale and complexity present unique challenges and opportunities, prompting researchers and practitioners to explore novel model training, optimization, and deployment methods. This literature review focuses on various techniques for reducing resource requirements and compressing large language models, including quantization, pruning, knowledge distillation, and architectural optimizations. The primary objective is to explore each method in-depth and highlight its unique challenges and practical applications. The discussed methods are categorized into a taxonomy that presents an overview of the optimization landscape and helps navigate it to understand the research trajectory better.
\end{abstract}


\section{Introduction}
In recent years, Large Language Models (LLMs) have emerged as the cornerstone of Natural Language Processing (NLP), revolutionizing various domains with unprecedented capabilities. These versatile models have demonstrated remarkable abilities in diverse applications, ranging from assisting in code generation \cite{Li2023StarCoderMT} \cite{Chen2021EvaluatingLL}, to facilitating news summarization \cite{Wei2022EmergentAO} \cite{Lewis2019BARTDS}, and even augmenting information retrieval systems for improved search accuracy and efficiency \cite{Lewis2020RetrievalAugmentedGF} \cite{DBLP:journals/corr/abs-2312-10997}.
Furthermore, these models' sheer scale and complexity present unique challenges and opportunities, prompting researchers and practitioners to explore novel model training, optimization, and deployment methods. Optimizing large models for speed, reducing resource consumption, and making them more accessible is a significant part of LLM research.

\begin{figure*}[h]
    \centering
    \includegraphics[scale=0.4]{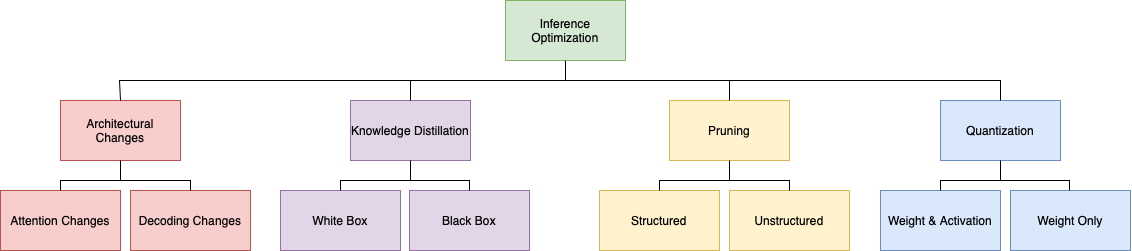}
    \caption{Taxonomy of optimization techniques}
    \label{fig:taxonomy}
\end{figure*}

The primary objective of this research paper is to explore various techniques for reducing resource requirements and compressing large language models, including analyzing each method in-depth and highlighting its unique challenges and practical implications. The discussed methods include quantization, pruning, knowledge distillation, and architectural optimizations. To better understand the relationship between these techniques, they are categorized into a taxonomy that presents an overview of the optimization landscape and helps navigate it for a better understanding of the research trajectory. Refer to figure \ref{fig:taxonomy} for a visual representation of the categorization and to the respective section for a more detailed look at the discussed literature in each category.

\section{Preliminaries}
\subsection{Transformers for Language Modeling}
In recent years, the transformer has become the primary architecture for Natural Language Processing (NLP) tasks \cite{Brown2020LanguageMA} \cite{open-llm-leaderboard} \cite{Fu2023ChainofThoughtHA} \cite{Radford2019LanguageMA}. The prominence is because of the attention mechanism, which enables the transformer to efficiently focus on different text parts and learn complex language structures \cite{Vaswani2017AttentionIA} \cite{Kim2023FullSO}. The mechanism aims to determine the significance of various elements in a sequence about a specific element \cite{Vaswani2017AttentionIA} \cite{Kim2023FullSO}.
These learned structures enable the transformer to solve complex NLP problems. In addition to that, the transformer typically comprises multiple transformer blocks, each containing an attention module and a feed-forward module \cite{Vaswani2017AttentionIA} \cite{Devlin2019BERTPO}. The feed-forward model facilitates the learning of mappings between the input and output and can be used in the encoder or decoder blocks \cite{Vaswani2017AttentionIA} \cite{Kim2023FullSO}. The encoder processes the input sequence in parallel, while the decoder blocks are autoregressive, performing inference once per output token \cite{Kim2023FullSO}.

\subsection{Emergent Abilities in Large Language Models}
In recent years, models have been adapted to different tasks. That means training a new model each time a different task is required, for example, summarization or question answering \cite{zhang2019pegasus} \cite{https://doi.org/10.48550/arxiv.2111.09645}. For example a fine-tuned version of RoBERTa \cite{Liu2019RoBERTaAR} for question answering trained on the SQuAD dataset \cite{Rajpurkar2016SQuAD1Q} or DistillBERT \cite{Sanh2019DistilBERTAD} for text classification on the SST-2 dataset \cite{socher-etal-2013-recursive}. Providing a different model can be challenging if a suitable one still needs training. However, large language models have special abilities that allow them to execute new, unseen tasks simply by understanding what to do by providing a task description with a few examples. These abilities are known as \textit{Emergent Abilities} \cite{Wei2022EmergentAO} \cite{Brown2020LanguageMA}. For example, an LLM can use these abilities to solve complex mathematical problems by only providing the necessary context in the prompt without retraining \cite{Wei2022ChainOT}. Therefore, addressing Emergent Abilities can be categorized by their prompting strategies. The authors of \cite{Wei2022EmergentAO} divide it into Few-Shot and Prompt Augmentation strategies. Few-shot prompting provides relevant examples to infer the desired behavior. At the same time, prompt augmentation captures all strategies that help the model better understand the desired behavior without explicitly providing few-shot examples. A prominent example would be chain-of-thought \cite{Wei2022ChainOT}, which provides the model with relevant intermediate steps to solve the problem and a task description.

\subsection{Hardware Representations of Numbers}
Graphic Processing Units (GPU) are used in machine learning to speed up computations \cite{Paszke2019PyTorchAI} \cite{Chetlur2014cuDNNEP} \cite{tensorflow}. To use the improvement, the hardware representation of numbers is in bits for the GPU to understand \cite{IEEEFP}. Typically, the representation uses 32 bits, the number of bits used to convey a number's meaning to the device. In machine learning, this is referred to as bit precision. Higher precision enables conveying more significant numbers, and lower precision means conveying fewer numbers. Reducing precision requires less memory space \cite{lin2020towards} \cite{wang2018training} \cite{lin2023awq}.

\begin{figure}[h]
    \centering
    \includegraphics[scale=0.35]{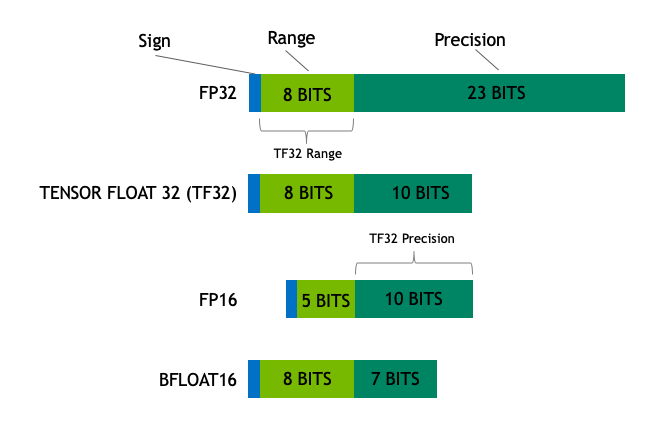}
    \caption{Different representation formats used in machine learning \cite{kharya2020}}
    \label{fig:bit_representation}
\end{figure}

Displaying a number on a hardware level involves three parts: the \textit{Sign}, the \textit{Range}, and \textit{Precision}, as shown in figure \ref{fig:bit_representation}. Whether a number is positive or negative is indicated by the Sign \cite{IEEEFP}. The Range refers to the amount of representable numbers. The larger the Range, the more numbers (higher and lower) can be represented \cite{IEEEFP}. Precision refers to how many decimals can be accurately represented \cite{IEEEFP}. The larger the Precision, the more fine-grained the numbers can be.

For LLMs, memory usage is determined by the amount of parameters and their precision. Due to the high memory requirements of FP32, LLMs default to using FP16. For a model with approximately 175 billion parameters, the required memory space would be approximately 334 GB.

\section{Inference Optimization}
Inference Optimization describes the procedure for enhancing the speed, efficiency, and performance of an LLM while preserving the quality of an uncompressed baseline. For large language models, this involves processing the input more rapidly and generating output more efficiently or with greater resource efficiency.

\subsection{Quantization}

Quantization involves converting high-precision numbers into a lower-precision space while preserving their meaning and significance \cite{shen2020q} \cite{Dettmers2022LLMint88M} \cite{zafrir2019q8bert}. The real advantage here is that the memory requirements are significantly lowered because fewer bytes are used to represent the number in a device like a GPU. This allows the device to load data more quickly and perform computations more efficiently, leading to tangible improvements in memory consumption, computational speed, and energy efficiency  \cite{jia2019dissecting} \cite{Kim2023FullSO} \cite{kuzmin2022fp8} \cite{lin2023awq} \cite{wu2023zeroquant}. These benefits enable more flexible deployment of LLMs on hardware-constrained devices, which is a practical application of this technique.

One way to achieve this is by converting the high-precision floating-point number into a lower-precision one. Another strategy reduces the precision while transforming the floating-point number into an integer \cite{Dettmers2022LLMint88M} \cite{frantar2022gptq} \cite{lin2023awq}. Therefore, the following paragraphs mainly discuss integer quantization in more detail and illuminate literature focusing on float-point numbers.

\begin{figure}[h]
    \centering
    \includegraphics[scale=1]{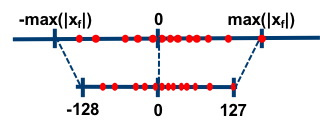}
    \caption{Symmetric Quantization example \cite{symAsym_intel_pic}}
    \label{fig:sym_absmax}
\end{figure}

Integer quantization can be divided into uniform and non-uniform strategies. These strategies describe how the transformation is performed. \textit{Uniform quantization} transforms the floating-point numbers with evenly spaced intervals and maps each interval to an integer value. \textit{Non-uniform quantization} may not have evenly spaced intervals \cite{Kim2023FullSO} \cite{shen2020q}. However, due to the difficulty of using non-uniform quantization with current accelerators such as GPUs, research primarily focuses on uniform quantization, as highlighted by \cite{Dettmers2022LLMint88M} \cite{Kim2023FullSO}.

\begin{figure}[h]
    \centering
    \includegraphics[scale=1]{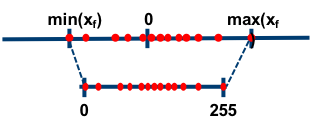}
    \caption{Asymmetric Quantization example \cite{symAsym_intel_pic}}
    \label{fig:asym_zerop}
\end{figure}

Uniform quantization is divided into symmetric and asymmetric quantization, which categorizes and describes how the transformation is performed. \textit{Symmetric Quantization} converts the floating point range in a way that sets the zero point of the floating-point range in the same place in the integer space, as visualized in the figure \ref{fig:sym_absmax} \cite{Dettmers2022LLMint88M} \cite{shen2020q} \cite{Yao2022ZeroQuantEA}. \textit{Asymmetric Quantization} transforms the minimum and maximum values of the floating-point range to be the lowest and highest value in the integer space, visualized in the figure \ref{fig:asym_zerop} \cite{Dettmers2022LLMint88M} \cite{shen2020q}.

\textbf{Appliance on the Transformer.} The reduction of precision can be applied to various aspects of the transformer. The two most common strategies are the quantization of weights only or weights and activations. Therefore, the reviewed literature gets categorized into quantization papers that discuss weight-only or weight-and-activation strategies. For a deeper look into the discussed literature, refer to table \ref{tab:quant_overview}.

There are different dimensions to the appliance of quantization to an LLM. For every quantization strategy, it is important to highlight which layer, which layer elements, and what bit precision the compression affects. For example, one of the first experiments focused on compressing the encoder layers of the BERT architecture while only focusing on weights \cite{Dettmers2022LLMint88M} \cite{lin2023awq} \cite{frantar2022gptq}. This evolved to experimenting with the quantization of weights and activations \cite{Yao2022ZeroQuantEA} \cite{Yao2023ZeroQuantV2EP}. This resulted in experimenting with different bit precision, which is currently focusing on 8bit \cite{Dettmers2022LLMint88M} \cite{Yao2022ZeroQuantEA}, 4bit precision\cite{lin2023awq} \cite{frantar2022gptq} or lower \cite{Lee2023OWQOW} \cite{Shao2023OmniQuantOC} \cite{Dettmers2023SpQRAS}. Therefore, these dimensions are highlighted in each piece of literature to provide a comprehensive image of the proposed method.
First, the weight-only literature is discussed, followed by literature discussing weight-and-activation quantization.

\begin{table}[htbp]
  \centering
  \caption{Literature Overview for Quantization}
  \label{tab:quant_overview}
  \begin{tabular}{cc}
    \toprule
    \multicolumn{2}{c}{\textbf{Quantization Literature}} \\
    \midrule
    Weight only & Weight \& Activation \\
    \midrule
    LLM.int8() \cite{Dettmers2022LLMint88M} & ZeroQuant \cite{Yao2022ZeroQuantEA} \\
    GPTQ \cite{frantar2022gptq} & ZeroQuantV2 \cite{Yao2023ZeroQuantV2EP} \\
    AWQ \cite{lin2023awq} & SmoothQuant \cite{Xiao2022SmoothQuantAA} \\
    OWQ \cite{Lee2023OWQOW} & \\
    SpQR \cite{Dettmers2023SpQRAS} & \\
    \bottomrule
  \end{tabular}
\end{table}

LLM.Int8()\cite{Dettmers2022LLMint88M} applies 8-bit weight quantization using a vector-wise transformation scheme. The study found that outlier dimensions have a significant impact when quantizing models beyond billions of parameters. They concluded that to achieve high-quality quantization, mixed-precision decomposition is needed, which keeps these outliers in a higher precision format. GTPQ \cite{frantar2022gptq} investigated the possibility of using 4-bit for compression. Their approach employs a layer-wise scheme, where a reconstruction problem is solved in each layer. This yielded a higher quality quantization by reducing transformation inaccuracies. Dettmers and Zettlemoyer \cite{Dettmers2022TheCF} researched the trade-off between quantization bit width and model size. They found that 4-bit compression is almost universally optimal for achieving performance improvements while maintaining quality. AWQ \cite{lin2023awq} focuses on linear layers and employs a per-channel quantization scheme. The activations in each channel are analyzed to identify salient weights and quantized differently to preserve the importance of these weights. However, all weights are assigned the same bit-width to avoid the performance cost of mixed-precision quantization. OWQ \cite{Lee2023OWQOW} focuses only on the linear layers and follows the same principle as AWQ by analyzing the importance of weights. They conclude with a different quantization scheme that keeps important weights in higher precision while compressing all others. SpQR \cite{Dettmers2023SpQRAS} is based on the insights of GPTQ and uses a layer-by-layer approach. The quantization process is divided into two steps. Firstly, weights are identified and isolated based on their output behavior. Secondly, a quantization step is applied to the remaining weights, reducing them to 3-4 bits.

In addition to weight-only quantization, quantization can also be applied to weights and activations. ZeroQuant \cite{Yao2022ZeroQuantEA} conducted research on the compression of weights and activations for large language models. They found that quantizing activations with a finer granularity than weights is essential to preserve quality. They conclude that weights allow 4-bit compression but keep quantized activations in higher 8-bit precision to preserve quality. They point out that for the BLOOM \cite{Scao2022BLOOMA1} family of models, better quantization could be achieved with a more coarse-grained granularity and higher precision rather than finer-grained granularity with lower precision. SmoothQuant \cite{Xiao2022SmoothQuantAA} focuses on the linear layers and uses a per-channel smoothing technique. They conclude shifting the quantization difficulty from the activations to the weights retains more activation information. OmniQuant \cite{Shao2023OmniQuantOC} optimizes weight and activation quantization by introducing separate learnable parameters for both weights and activations. This approach enables a wide range of quantization settings to be adjusted more effectively.


\textbf{The Practical Aspect.} The practical application of quantization can be divided into Quantization Aware Training (QAT) and Post Training Quantization (PTQ). QAT aims to improve the efficiency and accessibility of language models by applying quantization during training. With quantization employed during training, the model can now learn low-precision representations, and subsequent weight updates can help mitigate any potential quality degradation caused by data type transformation. The principle was first applied using small BERT models, demonstrating the quantization possibilities \cite{shen2020q} \cite{zafrir2019q8bert}. This enables the fine-tuning of LLMs with billions of parameters with only a handful of GPUs \cite{Dettmers2022LLMint88M} \cite{Kim2023MemoryEfficientFO}. Multiple LLMs can now efficiently be trained and swapped out for inference when needed.

If the aim is to train a model for a specific task while considering high resource requirements, Quantization Aware Training can be helpful. Nevertheless, many fine-tuned models are available, like llama2-chat \cite{Touvron2023Llama2O} or CodeLlama \cite{Rozire2023CodeLO}. Post-training quantization can help manage storage and deployment challenges without significant weight changes or extensive training efforts. Therefore, there is a significant focus on researching PTQ techniques with minimal impact on quality.

\textbf{The quantization error.} Quantization can introduce inaccuracies due to data type transformation, which can significantly impact the quality of a compressed model. To make PTQ feasible, the focus is on reducing these inaccuracies as much as possible while preserving the model characteristics and knowledge representation. The challenges of using quantization effectively are discussed in the following paragraph.

LLM.Int8() \cite{Dettmers2022LLMint88M} identified a challenge in applying quantization to transformers with more than a billion parameters: the emergence of outlier features in large models. These outliers are difficult to quantize due to their extreme magnitudes, which can cause a shift in integer representation. In addition, AWQ \cite{lin2023awq} found through an analysis of the resulting activations that not all weights are of equal importance.
ZeroQuant \cite{Yao2022ZeroQuantEA}, ZeroQuant-V2 \cite{Yao2023ZeroQuantV2EP}, and SmoothQuant \cite{Xiao2022SmoothQuantAA} found that quantizing activations is significantly more challenging and leads to a more significant degradation in quality. This conclusion is why the latter proposed shifting the complexity from the activations to the weights.

\textbf{Practical Constraints.} Practical challenges must be considered to reduce resource consumption and speed up computation. Many methods rely on hardware support for particular data types, like 8-bit integers and special operators, to use them efficiently \cite{Wu2020IntegerQF}. So, there are two practical dimensions to remember: The amount of resources needed to compute the compression and the inference constraints needed to achieve a practical speed-up.

The impact of bit width, quantization schema, and granularity on quantization efficiency is significant. For instance, ZeroQuant \cite{Yao2022ZeroQuantEA} and SpQR \cite{Dettmers2023SpQRAS} can compress models with around 175 billion parameters in a matter of hours, while other methods require more compute time to achieve the same level of compression \cite{Xiao2022SmoothQuantAA} \cite{Dettmers2022LLMint88M} \cite{shen2020q}. Therefore, it is essential to consider this factor when selecting a quantization variant.

The resource requirements during compression and the constraints in using the compressed model are essential for consideration. Each method applies its compression differently, which could result in different inference behaviors. Methods such as LLM.Int8() \cite{Dettmers2022LLMint88M} can effectively reduce the memory consumption of LLMs, but it does not achieve a significant computational speed-up because of the need for dequantization during inference. The cause is the proposed mixed-precision approach, which significantly impacts quantization and decomposition overhead and relies on optimized GPU kernels. Another constraint is the hardware support for the quantization target. Additional overhead will slow down the compressed model if the hardware does not support the quantization target. For example, ZeroQuant \cite{Yao2022ZeroQuantEA} provides a custom Nvidia GPU kernel named CUTLASS to improve integer multiplications and reduce data loading overhead. OWQ \cite{Lee2023OWQOW} and SpQR \cite{Dettmers2023SpQRAS} offer custom kernels and back-end implementations to accelerate data loading overhead and matrix multiplications.

Therefore, the targeted deployment environment significantly impacts the optimization technique chosen.
Furthermore, it is essential to consider that if methods rely on custom code for specific hardware, their deployment may be more complicated than expected. In AWQ \cite{lin2023awq}, the authors mention that they built the algorithm upon the default PyTorch \cite{Paszke2019PyTorchAI} API, which does not rely on custom code and makes the use of a wide range of hardware configurations possible.

\textbf{In Conlusion.} The choice of quantization method depends on several factors, including resource requirements, inference constraints, and the desired balance between compression efficiency and model quality.
\begin{itemize}
  \item Quantization involves representing numerical data in a format that can be understood by hardware, such as GPUs.
  \item The main concept is converting precise numerical values into less precise ones while retaining their meaning, thus reducing memory and computation complexity. 
  \item Quantization can be divided into uniform and non-uniform. Uniform quantization evenly spaces intervals of floating point numbers.
  \item The uniform quantization, whether symmetric or asymmetric, can affect the data distribution mapping.
  \item Quantization can be applied either during training, known as Quantization Aware Training (QAT), or after training, known as Post Training Quantization (PTQ).
  \item Inaccuracies may arise from data type transformations, which can significantly impact model quality.
  \item Many different methods focus on quantizing weights, activations, or both. They use different quantization schemes and bit widths.
  \item Resource requirements for optimization and inference constraints must be considered.
  \item Constraints for quantization include quantization overhead, hardware support for the quantization target, and deployment complexity.
\end{itemize}

\subsection{Pruning}
Pruning describes identifying and removing redundant weights while trying to retain the properties of the uncompressed baseline \cite{LeCun1989OptimalBD} \cite{Frantar2023SparseGPTML}. There are two types of pruning: structured and unstructured pruning. \textit{Structured Pruning} focuses on preserving the original structure of the network by only removing high granularity structures like rows and columns, connections, and hierarchical structures \cite{Zhang2023LoRAPrunePM} \cite{Santacroce2023WhatMI} \cite{Ma2023LLMPrunerOT}. \textit{Unstructured Pruning} can remove single weights, resulting in an irregular sparse structure \cite{Frantar2023SparseGPTML} \cite{syed2023prune} \cite{Sun2023ASA}. The investigated literature can be viewed in table \ref{tab:pruning_overview}.

\begin{table}[htbp]
  \centering
  \caption{Literature Overview for Pruning}
  \label{tab:pruning_overview}
  \begin{tabular}{cc}
    \toprule
    \multicolumn{2}{c}{\textbf{Pruning Literature}} \\
    \midrule
    Structured Pruning & Unstructured Pruning \\
    \midrule
    LoRaPrune \cite{Zhang2023LoRAPrunePM} & SparseGPT \cite{Frantar2023SparseGPTML} \\
    What Matters in Structured Pruning? \cite{Santacroce2023WhatMI} & Prune and Tune \cite{syed2023prune}  \\
    LLM-Pruner \cite{Ma2023LLMPrunerOT} & Wanda \cite{Sun2023ASA} \\
    \bottomrule
  \end{tabular}
\end{table}

Several studies are researching unstructured pruning. Frantar \cite{Frantar2023SparseGPTML} proposed SparseGPT, which focuses on weight matrices and pruning individual weights. The approach aims to keep quality degradation in check by rephrasing the pruning problem as a sparse regression. This results in the pruning of current weights while adjusting not-yet-pruned weights. Prune and Tune \cite{syed2023prune} improves upon SparseGPT by introducing fine-tuning at each pruning step. The authors of \cite{Sun2023ASA} do not rely on retraining or iterative weight updates. This approach is similar to LLM.Int8() \cite{Dettmers2022LLMint88M}, by calculating the importance of weights based on their activations and subsequent pruning of unimportant weights.

In addition to structured pruning, several studies investigate the appliance of unstructured pruning. LoRAPrune aims to increase pruning efficiency by using LoRA mechanisms to estimate weight importance. \cite{Zhang2023LoRAPrunePM}. Using structured pruning and LoRA mechanisms for importance estimation enables the pruned model to merge fine-tuned LoRa weights, which improves efficiency. The calculation of weight importance is also a major factor of \cite{Santacroce2023WhatMI} following a similar strategy to LLM.Int8() \cite{Dettmers2022LLMint88M} and AWQ \cite{lin2023awq}. They propose identifying prunable weights by calculating two key figures: Sensitivity and Uniqueness. Sensitivity looks at a neuron's output and provides insights into its importance. Uniqueness provides insights into the variance, by telling how much a neuron differs from others. A low uniqueness indicates redundancy, and other neurons could reconstruct the output. Using these key figures in combination identifies prunable structures. LLM-Pruner \cite{Ma2023LLMPrunerOT} proposes a three-step process: (1) The discovery stage, where relevant structures and their dependencies get identified. (2) The estimation stage is where the identified structures get valued, and a decision is made on which structures are pruneable. (3) In the recovery stage, the error from pruning is mitigated via an efficient training step.

\textbf{Challenges of pruning.} Previously pruning required a retraining step to maintain the quality and characteristics of the uncompressed baseline \cite{Ma2023LLMPrunerOT} \cite{Dettmers2023QLoRAEF} \cite{Frantar2023SparseGPTML}. However, recent research has attempted zero-shot pruning without significant retraining efforts \cite{Ma2023LLMPrunerOT} \cite{Sun2023ASA}. Identifying all relevant structures for pruning is challenging, resulting in either a falsely or overpruned worse model or missing potential optimizations \cite{Frantar2023SparseGPTML} \cite{Ma2023LLMPrunerOT} \cite{Sun2023ASA}.

\textbf{Practical constraints of pruning.} Unstructured and structured pruning of models with billions of parameters can be challenging to achieve efficiently due to the vast number of possible structures available for pruning. Finding suitable structures for pruning can be challenging and computationally intensive \cite{Frantar2023SparseGPTML}. For instance, Frantar et al. \cite{Frantar2023SparseGPTML} utilizes insights from GPTQ \cite{frantar2022gptq} to prune a GPT-style model with 175 billion parameters in approximately four hours. However, Sun et al. \cite{Sun2023ASA} improved the efficiency of the prune-metric to achieve faster compression, allowing for real-time pruning if needed.
Unstructured pruning presents its own set of challenges. For example, pruned models are no longer compatible with LoRA-weights due to the irregular structures that remain after compression \cite{Zhang2023LoRAPrunePM}. This constraint could impede specific deployment scenarios where task-specific models are necessary and real-time pruning is applied. 

It is essential to consider the constraints of using pruned models for inference. To achieve a speed-up, the hardware in the production environment has to support sparse models. LoRA-Prune \cite{Zhang2023LoRAPrunePM} emphasizes that specific unique hardware configurations are needed for unstructured pruning to utilize the improvement. SpraseGPT \cite{Frantar2023SparseGPTML} mention specific hardware support for structured pruning for their performance evaluation. Therefore, identifying if the targeted hardware supports models structured or unstructured pruned models is important.

\textbf{In Conclusion.}
\begin{itemize}
    \item Pruning involves identifying and removing redundant weights while maintaining the properties of the uncompressed baseline.
    \item Two types of pruning: structured and unstructured pruning.
    \item Unstructured Pruning focuses on removing individual weights, resulting in an irregular sparse structure.
    \item Structured Pruning preserves the original structure of the network by removing high-granularity structures like rows, columns, and connections.
    \item Pruning requires retraining to maintain quality and characteristics, but recent research attempts zero-shot pruning without significant retraining.
    \item Identifying all relevant structures for pruning can be challenging, impacting the effectiveness of zero-shot pruning methods.
    \item Hardware constraints during inference, sparse model support, and pruning ratios significantly impact efficiency.
\end{itemize}

\subsection{Knowledge Distillation}
Knowledge Distillation describes a fine-tuning and compression technique that allows the transfer of knowledge of a large complex model into a smaller streamlined and efficient model \cite{Zhao2022DecoupledKD} \cite{Breiman1996BORNAT} \cite{Bucila2006ModelC} \cite{Ba2013DoDN}. Transferring knowledge is done by training a smaller student model, with the outputs of a bigger teacher model \cite{Gu2023KnowledgeDO}. There are two distinct approaches, \textit{Black-Box Knowledge Distillation}, where only the outputs of the teacher model are available \cite{Huang2022IncontextLD} \cite{LI2022ExplanationsFL} \cite{Fu2023SpecializingSL} \cite{Wu2023LaMiniLMAD} and \textit{White-Box Knowledge Distillation} where also the teacher model parameters are utilized \cite{Gu2023KnowledgeDO} \cite{Agarwal2023OnPolicyDO} \cite{Jha2023HowTT}. The detailed listing of all reviewed literature can be seen in table \ref{tab:kd_overview}

\begin{table*}[htbp]
  \centering
  \caption{Literature Overview for Knowledge Distillation}
  \label{tab:kd_overview}
  \small
  \begin{tabular}{p{0.45\columnwidth} p{0.45\columnwidth}} 
    \toprule
    \multicolumn{2}{c}{\textbf{Knowledge Distillation Literature}} \\
    \midrule
    \textbf{White-Box KD} & \textbf{Black-Box KD} \\
    \midrule
    MiniLLM \cite{Gu2023KnowledgeDO} & In-context Learning Distillation \cite{Huang2022IncontextLD} \\
    On-Policy Distillation of LMs \cite{Agarwal2023OnPolicyDO} & Explanations from LLMs Make Small Models Better \cite{Magister2022TeachingSL} \\
    How To Train Your (Compressed) LLM \cite{Jha2023HowTT} & LLMs Are Reasoning Teachers \cite{Ho2022LargeLM}\\
     & Specializing Smaller Models  \cite{Fu2023SpecializingSL}\\
     & Distilling Step-by-Step! \cite{Hsieh2023DistillingSO}\\
     & Distilling Reasoning Capabilities Into Smaller Models \cite{Shridhar2022DistillingRC}\\
     & SCOTT \cite{Wang2023SCOTTSC}\\
     & PaD: Program-aided Distillation \cite{Zhu2023PaDPD}\\
     & Can LMs Teach Weaker Agents? \cite{Saha2023CanLM}\\
     & Lion: Adversarial Distillation \cite{Jiang2023LionAD}\\
     & LaMini-LM \cite{Wu2023LaMiniLMAD} \\
    \bottomrule
  \end{tabular}
\end{table*}

\textbf{White-Box Knowledge Distillation.} Utilizing the teacher model's internal workings to improve the student model's learning is called White-Box Knowledge Distillation. The core idea is to use the teacher's distribution and parameter settings to enable the student to learn more effectively. The student model's increased quality is achieved by reducing the \textit{Kullback-Leibler Divergence} (KLD). \cite{Gu2023KnowledgeDO} \cite{Agarwal2023OnPolicyDO} \cite{Jha2023HowTT}

The authors of MiniLLM \cite{Gu2023KnowledgeDO} aim to minimize distribution differences between the teacher and student but point out that this can lead to drifts due to the vast output space of LLMs, which the student model may not be able to replicate. To mitigate this, they propose to optimize the inverse KLD, which encourages the student to learn probabilities that prioritize correctness. Agarwal et al. \cite{Agarwal2023OnPolicyDO} investigate the distribution behavior of both models during distillation and result in two fundamental problems: (1) distribution mismatch between the teacher and the student, and (2) the student's lack of expressive power to imitate the teacher effectively.  To achieve more precise imitation, they sampled the outputs of the student during training to improve learning. They also researched under-specification, where optimizing KDL was proposed. The authors of \cite{Jha2023HowTT} criticize that the inherent principle of Knowledge Distillation poses a problem, as previous strategies focus on a specific task setting instead of transferring broad skills present in today's LLMs. To achieve task-agnostic distillation, they propose truncating the larger model and using its layers to initialize the smaller model. The smaller model is then distilled using the language modeling objective.

\textbf{Black-box Knowledge Distillation.} Black-Box Knowledge Distillation differs in that only a teacher's response is accessible, and utilizing parts of a teacher model is no longer possible \cite{Huang2022IncontextLD} \cite{LI2022ExplanationsFL} \cite{Wu2023LaMiniLMAD} \cite{Zhu2023PaDPD}. To effectively use Black-Box Knowledge Distillation, the learning process is augmented with recent research on emergent abilities, which enables the usage of methods like Chain-of-Thought \cite{Fu2023ChainofThoughtHA} to improve quality. This results in using various prompting techniques to support the Distillation of the larger teacher model into smaller, more efficient one.

One of those prompting techniques is In-Context Learning, which uses a natural language prompt in combination with relevant descriptions and examples to enable a model to fulfill or solve a task. This technique distills smaller models to transfer in-context few-shot and language modeling abilities from the teacher model. There are two distilling strategies to transfer in-context learning: Meta In-Context Tuning (Meta-ICT) and Multitask In-context Tuning (Multitask-ICT) \cite{Huang2022IncontextLD}. Meta-ICT trains the smaller model in the \textit{Pre-Training} stage with examples of in-context learning on diverse tasks. The result is a fine-tuned model that learns to identify different tasks and uses this ability to transfer this knowledge to unseen tasks during the \textit{Task Adaption} stage. Multitaks-ICT, proposed by  \cite{Huang2022IncontextLD}, instead of providing the adjustment in the Pre-Training stage, the adjustment happens during the Task-adaption stage. The results show that Multitask-ICT is more precise and has an advantage over Meta-ICT.

The following strategy for distilling uses Chain-of-Thought to help the student learn the abilities of the teacher model. Augmenting the prompt with step-by-step instructions on finding a solution is the distinction from in-context tuning, which only provides task descriptions\cite{Wei2022ChainOT}. The authors of \cite{LI2022ExplanationsFL} researched prompt augmenting and found that including instructions on how to infer a solution significantly helps the student at task adaption. The authors of \cite{Magister2022TeachingSL} adjusted these insights for Knowledge Distillation and divided the fine-tuning phase into steps. Step one is augmenting existing instruction tuning datasets with teacher-generated Chain-of-Thought explanations. The student then learns via these augmented datasets by mimicking the teacher-generated explanations. Providing intermediate steps to find a solution and rationales for why a solution is correct further enhances the learning of the student model \cite{Ho2022LargeLM}.
The authors further argue that providing more than one rationale helps the student model more effectively because complex problems can have multiple solutions. The quality of the student model is dependent on the quality of the used dataset for distilling. Not Evaluating abilities, not part of the dataset used for distilling, is recognized as a potential problem by \cite{Fu2023SpecializingSL}. They argue that focusing a student model on a specific set of tasks makes it a specialized model instead and incapable of generic reasoning. The authors of \cite{Hsieh2023DistillingSO} shift the paradigm of distilling from using the question in combination with the rationale for learning to predict the answer with a fitting rationale. They leverage an unlabeled dataset and generate labels with rationales as justifications. Labels and rationales are then used to distill smaller models. Dividing the distillation problem into understanding problem structures and providing solution steps for these sub-problems, is proposed by \cite{Shridhar2022DistillingRC}. This division results in training two models, where the first model learns to decompose the problem into smaller sub-problems and annotates the prompt with sub-questions. The second model learns to use these sub-questions to generate a higher-quality answer. Using Chain-of-thought in combination with rational generation is identified as problematic by \cite{Wang2023SCOTTSC}. There are two fundamental problems: LLMs are prone to hallucination and could provide false rationals, which the student learns. Also, the student may take shortcuts and generate an answer independently from the provided rationale. To combat these problems, they proposed contrastive decoding to ground an answer to the rational. This decoding process means the student has to use counterfactual reasoning, which results in it being more truthful. The authors of \cite{Zhu2023PaDPD} also focus on truthfulness by generating rationals in the form of small Python programs. These are then automatically evaluated, and faulty reasoning is more easily discovered. Another approach to increase the quality is to provide feedback to the student \cite{Saha2023CanLM}. The teacher model uses the generations of the student and decides to intervene when the student is struggling. The intervention is implemented with a Theory of Mind approach. This means the teacher creates a small mental model of the student for the intervention decision.

Another strategy for distilling would be to use only detailed descriptions of instructions without examples or descriptions of intermediate steps. To enable the smaller model to solve unseen tasks, it gets trained on a large number of different instructions \cite{Ouyang2022TrainingLM} \cite{Brooks2022InstructPix2PixLT}. Lion \cite{Jiang2023LionAD} divides the distilling process into three steps. In the \textit{Imitation Stage}, the student responses are aligned to the teacher's. The teacher then identifies false answers to gather complicated instructions in the subsequent \textit{Discrimination Stage}. The final, \textit{Generation Stage} uses gathered instructions and enriches them with more details to help the student learn more effectively. La-Mini-LM \cite{Wu2023LaMiniLMAD} investigated the current instruction following distillation strategies. They found a lack of available small-scale distilling datasets, and available datasets need to provide more diversity to produce a high-quality model.
Moreover, student evaluation focuses too narrowly on one model family, and the evaluation process needs to be more precise. To improve dataset diversity, a new distilling dataset is proposed. Furthermore, trained models on this dataset are thoroughly evaluated using a comprehensive testing methodology.

\textbf{Challenges.} There are significant challenges that impact distillation quality. In order to achieve more students with the teacher's abilities, the reduction of the distribution mismatch is actively investigated \cite{Gu2023KnowledgeDO} \cite{Agarwal2023OnPolicyDO}. The difficulty in reduction could stem from the lack of publicly available diverse distilling datasets \cite{Wu2023LaMiniLMAD}. This results in either manually creating or augmenting existing datasets. However, the diversity and detail of current datasets could also be a challenge in distilling capable students \cite{Wu2023LaMiniLMAD}.
Also, reaching specific quality aspects for the student is challenging. The authors of \cite{Agarwal2023OnPolicyDO} found that it is challenging distilling a student that has similar distribution and expressive power to the teacher. One possible reason investigated \cite{Gudibande2023TheFP}, which criticized that the student can only be as good as the used teacher. That means that the student learns hallucinations, bias, and toxicity behavior already present in the teacher. If only Black-Box Knowledge Distillation is available, the risk of only imitating the teacher's style without actually learning the behavior is present \cite{Gudibande2023TheFP}.

\textbf{Practical Constraints.} In many knowledge distillation strategies, a student-teacher setup is deployed. This results in two models loaded into memory during white-box and local black-box distillation, which requires a memory-potent environment \cite{Gu2023KnowledgeDO} \cite{Agarwal2023OnPolicyDO} \cite{Jha2023HowTT}. Running two models is costly because computing and memory are expensive. These costs also apply to APIs, where memory costs are factored into API usage costs. Therefore, it has to be considered if fine-tuning is needed given the resource requirements and if it is preferred over simple optimization techniques like quantization.

\textbf{In Conclusion.}
\begin{itemize}
    \item Knowledge distillation is a technique for fine-tuning and compressing models, with the goal of transferring knowledge from larger models to smaller ones.
    \item This entails training a smaller \say{student} model with the outputs of a larger \say{teacher} model.
    \item Two approaches, black-box and white-box knowledge distillation, are used.
    \item White-box knowledge distillation enables the student to utilize the teacher's parameters for a better understanding.
    \item Black-box knowledge distillation relies solely on the teacher's prediction.
    \item The standard student-teacher configuration in a white-box setting necessitates both models to operate in memory, which can be computationally demanding.
    \item Distributional imbalances, lack of diverse data sets, and imitation of undesirable teacher behaviors are among the quality challenges in knowledge distillation.
    \item Challenges also arise from the nature of black-box knowledge distillation, where students may mimic the teacher's style without understanding the desired behavior.
\end{itemize}

\subsection{Architectural Optimization}
The transformer performs well because of the attention mechanism. The computation of this mechanism requires much memory because of the key-value cache. During generations, predicting the next token requires the caching of all previous tokens in order to calculate the joint probabilities \cite{Kim2023FullSO} \cite{Kwon2023EfficientMM}. Therefore, one big focus is making the decoding process or the attention mechanism more efficient \cite{Beltagy2020LongformerTL}. The following discusses popular optimization strategies for optimizing the decoding phase.

\textbf{Paged Attention.} The authors of vLLM \cite{Kwon2023EfficientMM} focused on making the key-value cache more memory efficient. This optimization is achieved by transferring a memory-storing concept of operating systems and applying it to the attention mechanism. Here, the key-value cache is divided into blocks allocated efficiently in memory. The allocation is done by placing the blocks in non-contiguous parts of physical memory, which is more efficient.

\textbf{Windowed Attention.} Windowed attention, proposed by \cite{Beltagy2020LongformerTL}, tries to mitigate huge computational requirements with large sequence transformers. The memory requirements scale quadratically with sequence length due to the key-value cache \cite{Beltagy2020LongformerTL}. Windowed Attention uses sliding windows to focus attention more locally, reducing complexity. This focus is beneficial for masked language modeling, where only a local focus is needed, but it is counterproductive for classification, which needs a broader focus. For mitigation, the authors propose sliding windows with an augmented global window, which helps to broaden the context.

\textbf{Attention Sinks.} There are challenges when using sliding window attention. The model could collapse when the sliding window exceeds the cache size, and costly recomputation has to be performed \cite{Xiao2023EfficientSL}. The authors found that the first tokens get significant attention scores. Therefore, they concluded that instead of discarding the first tokens after the sliding windows further progress, a couple of the first tokens are kept. This results in quality and performance improvements, as stated in \cite{Xiao2023EfficientSL}.

\textbf{Flash Attention.} Not only is the complexity of the attention algorithm a potential bottleneck, but the device data management to fill it with data could also be problematic \cite{Dao2022FlashAttentionFA}. The goal of Flash-Attention \cite{Dao2022FlashAttentionFA} is to minimize the need for data movement while computing the attention matrices. They achieve this by regrouping operations from different parts of the LLM and keeping relevant data in memory to reduce memory loading overhead.

\textbf{Speculative Decoding.} The decoding phase of a decoder-only transformer is costly because of the sequential properties of generation \cite{Kwon2023EfficientMM} \cite{Dao2022FlashAttentionFA} \cite{Leviathan2022FastIF}. Instead of optimizing attention, Speculative Decoding \cite{Leviathan2022FastIF} focuses on making the decoding process more efficient. It operates under the principle of performing generations in parallel while concurrently verifying their necessity. The concurrency is achieved by using two models: one primary model and one support model. The support model provides token suggestions that the primary model accepts or rejects.
Speculative decoding significantly reduces inference time and resource consumption by maximizing the probability of acceptance for these speculative tasks while ensuring their outputs maintain the same distribution as those from the target model alone. The authors mention the effectiveness of speculative decoding across various tasks and model sizes, including unconditional generation, translation, summarization, and dialogue tasks, resulting in a notable 2x-3x latency improvement without impacting output quality.

\begin{figure}[h]
    \centering
    \includegraphics[scale=0.23]{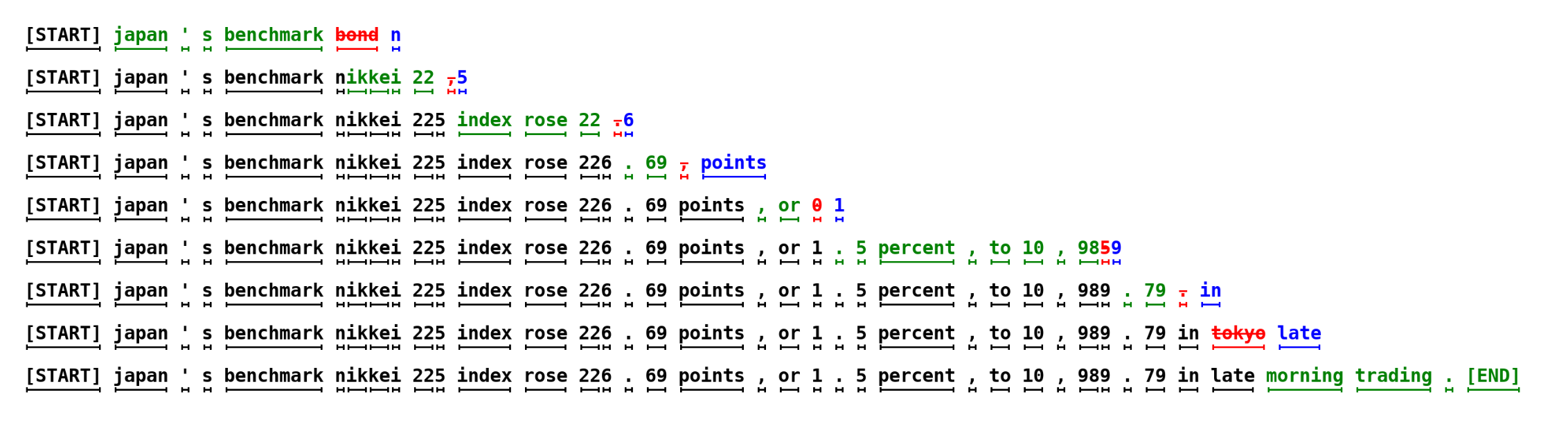}
    \caption{Example generation out of \cite{Leviathan2022FastIF}, where green are accepted generations, red and blue are rejections and corrections, respectively.}
    \label{fig:speculative decoding}
\end{figure}

\textbf{In Conclusion.}
\begin{itemize}
    \item The transformers' attention mechanism requires significant computational resources.
    \item To optimize the key-value cache, it can be divided into blocks that are stored in non-contiguous parts of physical memory.
    \item As the length of a transformer sequence increases, the complexity of attention increases exponentially.
    \item Sliding windows and attention sinks can help reduce this complexity.
    \item Flash Attention minimized expensive data movement by regrouping operations and keeping relevant data in memory.
    \item Speculative Decoding samples generations from a more efficient model and achieves its speed up by reducing the needed runs from the larger model.
\end{itemize}

\section{Conclusion}

This literature review explored multiple optimization techniques, their challenges, and practical applicability. It aims to provide insights into the current state of research and its practical considerations. The explored methods included quantization, pruning, knowledge distillation, and optimizations of the attention mechanism and decoding process. There are different grades of optimization development research and applicability. This grading makes categorizing these techniques into their current state of research and applicability possible. 
Compared to quantization, Pruning methods have to develop more precise methods of finding all available structures for compression. In contrast, quantization literature thoroughly discusses the implications of applying it to different LLM elements. The advancement in research leads to quantization, which mainly focuses on weights and tries to push compression via bit width. Knowledge distillation provides the most flexibility for affected abilities by choosing the distilling dataset and the prompting technique. It needs more training, which makes setup and resource requirements costly. In conclusion, the chosen optimization technique is dependent on the optimization goals, available resources, and target environment.

\bibliography{Quellen.bib}
\bibliographystyle{IEEEtran}

\end{document}